\documentclass{article}
\usepackage{algorithmicx}
\usepackage{tikz}
\usetikzlibrary{arrows.meta}
\usepackage{amsmath}

\usepackage{algorithm}
\usepackage{algpseudocode}


\usepackage{float}
\usepackage{subcaption}

\usepackage{graphicx}

 \usepackage[main, final]{neurips_2025}

\usepackage[utf8]{inputenc} 
\usepackage[T1]{fontenc}    
\usepackage{hyperref}       
\usepackage{url}            
\usepackage{booktabs}       
\usepackage{amsfonts}       
\usepackage{nicefrac}       
\usepackage{microtype}      
\usepackage{xcolor}         

\title{SOFA-FL: Self-Organizing Hierarchical Federated Learning with Adaptive Clustered Data Sharing}

\author{
  Yi Ni\\
  Carnegie Mellon University\\
  \texttt{\small yini@andrew.cmu.edu} \\
\And  
  Xinkun Wang\\
  Carnegie Mellon University\\
  \texttt{\small xinkunw@andrew.cmu.edu} \\  
\And
Han Zhang\\
  Carnegie Mellon University\\
  \texttt{\small hanz5@andrew.cmu.edu} \\
}

\begin{document}

\maketitle
\footnotetext{%
\textbf{Code repository:} Implementation available at 
\href{https://github.com/T0nyN1/SOFA-FL}{SOFA-FL GitHub repository}.%
}
\section{Introduction}
Federated Learning (FL) has emerged as a powerful paradigm that enables distributed model training and provides privacy preservation for large-scale decentralized systems, which has been one of the essential topics in distributed machine learning for its promising applicability \citep{YURDEM2024e38137, electronics14132512}. However, traditional FL faces a series of challenges, including data heterogeneity, limited personalization capabilities, and inflexibility in evolving environments \citep{Saeed2025}. Specifically, clients within FL systems usually struggle to deal with heterogeneous and temporally drifting data, leading to suboptimal global convergence and degraded performance on local tasks.

To address these issues, Hierarchical Federated Learning (HFL) has been proposed. HFL introduces multi-level structures that organize the clients into subgroups, which aggregates similar clients and allows for personalization. However, most HFL architectures rely on a fixed hierarchical topology or predetermined cluster numbers, which cannot effectively adapt to changes in clients’ data distributions. Consequently, they cannot effectively capture the temporally dynamic relationships among clients and manifest inherent defects for continual learning \citep{fang2024hierarchicalfederatedlearningmultitimescale, hamedi2025federatedcontinuallearningconcepts}.

\section{Methodology}
To overcome these limitations, we propose SOFA-FL (Self-Organizing Hierarchical Federated Learning with Adaptive Clustered Data Sharing), a framework that allows hierarchical federated systems to self-organize and self-adapt over time.The framework is composed of three parts: clustering using Dynamic Multi-branch Agglomerative Clustering (DMAC) algorithm, adjusting the architecture using Self-organizing Hierarchical Adaptive Propagation and Evolution (SHAPE) algorithm, and finally, sharing partial data between clients from upward gathering and downward distributing. 

We started the training with DMAC algorithm to construct an initial clustering structure and allow for clustering tree to have multiple branches to reduce the tree level. Then we started local update and global communication. After each round, we used Self-organizing SHAPE algorithm to modify the tree structure to allow for architecture evolution and adaptation to changing data distribution. By avoiding regrouping nodes each time, we can effectively reduce the time complexity and improve the robustness in the training phase. Finally, we added PDS within the clusters to alleviate the data heterogeneity and improve the test accuracy of clients as well as the cluster nodes. 

\subsection{Dynamic Multi-branch Agglomerative Clustering (DMAC)}

In SOFA-FL, clients self-organize into a tree-structured hierarchical topology by being dynamically grouped into clusters using hierarchical clustering based on latent representation distance or model gradient similarity. This hierarchical clustering structure also naturally aligns with real-world multi-level organizations, such as province–city–district hierarchies, enabling the model to capture personalization across different hierarchical levels. After several updates, the architecture will be updated using SHAPE algorithm to adjust the position of clients while persevering the overall structure for robustness. This mechanism enables the system to continuously adjust its topology, resulting in enhanced personalization capability and adaptive learning performance under nonstationary environments.

During the initialization stage, we will do warm-up local update and apply DMAC algorithm to cluster the clients based on their model weights from warm-up stage. The DMAC algorithm makes modifications upon traditional hierarchical clustering algorithm. Instead of merging two closest clusters together in each round, we set up a threshold and merge any clusters whose distance is less than the threshold, allowing multiple clusters to be merge at once, thus reducing the level of the clustering tree.

\begin{figure}[H]
    \centering
    \includegraphics[width=0.5\linewidth]{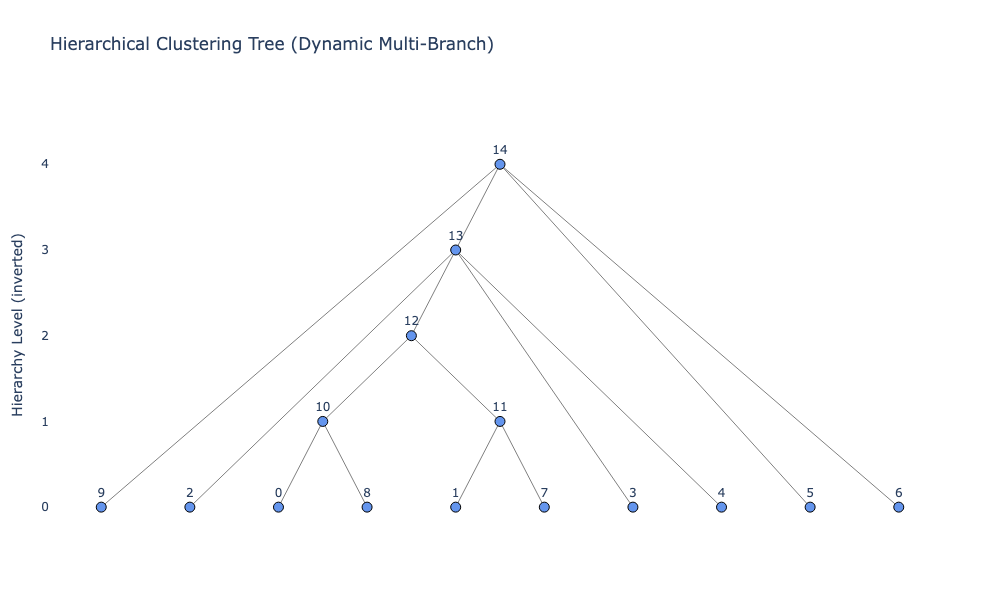}
    \caption{Diagram of Dynamic Multi-branch Agglomerative Clustering}
    \label{fig:DMAC}
\end{figure}


From Figure \ref{fig:DMAC}, the nodes at Level 4 represent the initial trainable clients in a real-world federated environment (e.g., users’ mobile devices). As we move upward in the hierarchy, each higher-level node corresponds to an aggregated model from its successors. 


\subsection{Self-organizing Hierarchical Adaptive Propagation and Evolution (SHAPE)}

The SHAPE mechanism dynamically restructures the hierarchical topology $\mathcal{T}$ through four atomic operations to adapt to concept drift and data heterogeneity. Grafting relocates a node to a closer parent when the distance improvement exceeds a tolerance threshold $\epsilon$. Pruning removes redundant intermediate nodes with only one child by directly connecting the child to its grandparent, thereby flattening the hierarchy. Consolidation merges sibling nodes whose distance $d(n_a, n_b) < \tau$ into a single node, preventing cluster fragmentation. Purification splits heterogeneous clusters using K-Means, accepting the partition only when all sub-clusters satisfy the incoherence threshold $\theta_{split}$. Figure \ref{fig:shape} illustrates these operations, with detailed pseudocode provided in Appendix~\ref{appendix:shape_algorithm}.

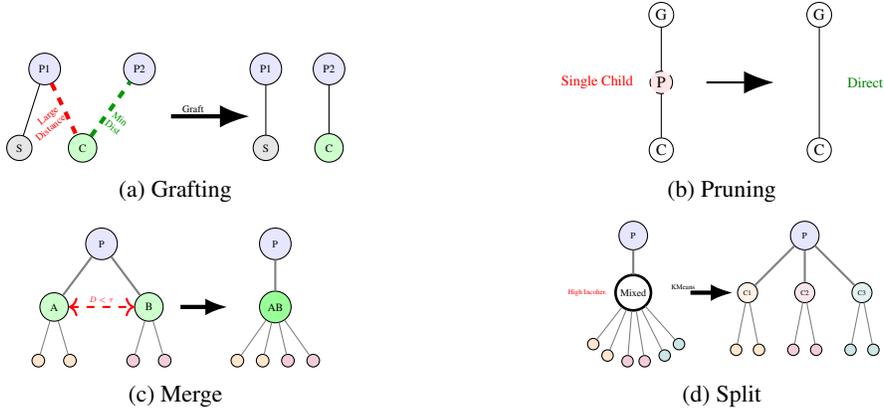
\begin{figure}[H] 
    \centering
    \begin{subfigure}[b]{0.48\linewidth}
        \centering
        \begin{tikzpicture}[scale=0.42, transform shape]
            \node[circle, draw, fill=blue!10, minimum size=1cm] (P1) at (0, 4) {P1};
            \node[circle, draw, fill=blue!10, minimum size=1cm] (P2) at (3, 4) {P2};
            \node[circle, draw, fill=gray!20, minimum size=0.8cm] (S) at (-0.8, 1.5) {S};
            \node[circle, draw, fill=green!20, minimum size=0.9cm] (C) at (1.2, 1.5) {C};
            \draw (P1) -- (S);
            \draw[ultra thick, dashed, red] (P1) -- node[left, rotate=40, font=\small, align=right] {Large\\Distance} (C);
            \draw[ultra thick, dashed, green!60!black] (C) -- node[right, font=\small, pos=0.4, rotate=-40, align=left] {Min\\Dist} (P2);
            \draw[-{Latex[length=4mm]}, ultra thick] (4, 2.5) -- (6.5, 2.5);
            \node[above, font=\small] at (4.7, 2.5) {Graft};
            \node[circle, draw, fill=blue!10, minimum size=1cm] (nP1) at (7, 4) {P1};
            \node[circle, draw, fill=blue!10, minimum size=1cm] (nP2) at (9, 4) {P2};
            \node[circle, draw, fill=gray!20, minimum size=0.8cm] (nS) at (7, 1.5) {S};
            \node[circle, draw, fill=green!20, minimum size=0.9cm] (nC) at (9, 1.5) {C};
            \draw (nP1) -- (nS);
            \draw (nP2) -- (nC);
        \end{tikzpicture}
        \caption{Grafting} 
        \label{fig:graft}
    \end{subfigure}
    \hfill 
    \begin{subfigure}[b]{0.48\linewidth}
        \centering
        \begin{tikzpicture}[scale=0.6, transform shape]
            \node[circle, draw, inner sep=2pt] (R1) at (0, 3) {G};
            \node[circle, draw, dashed, fill=red!10, inner sep=2pt] (M1) at (0, 1.5) {P};
            \node[circle, draw, inner sep=2pt] (L1) at (0, 0) {C};
            \draw (R1) -- (M1) -- (L1);
            \node[left, font=\small, red] at (-0.5, 1.5) {Single Child};
            \draw[-{Latex[length=4mm]}, thick] (1, 1.5) -- (2.5, 1.5);
            \node[circle, draw, inner sep=2pt] (R2) at (3.5, 3) {G};
            \node[circle, draw, inner sep=2pt] (L2) at (3.5, 0) {C};
            \draw (R2) -- (L2);
            \node[right, font=\small, green!50!black] at (4.0, 1.5) {Direct};
        \end{tikzpicture}
        \caption{Pruning}
        \label{fig:trim}
    \end{subfigure}
    
    \vspace{0.5em} 

    \begin{subfigure}[b]{0.48\linewidth}
        \centering
        \begin{tikzpicture}[scale=0.42, transform shape]
            \node[circle, draw, fill=blue!10, minimum size=1cm, font=\small] (P) at (2, 4.5) {P};
            \node[circle, draw, fill=green!20, minimum size=0.9cm] (A) at (0.5, 2.5) {A};
            \node[circle, draw, fill=green!20, minimum size=0.9cm] (B) at (3.5, 2.5) {B};
            \node[circle, draw, fill=orange!20, minimum size=0.4cm] (a1) at (0, 0.8) {};
            \node[circle, draw, fill=orange!20, minimum size=0.4cm] (a2) at (1, 0.8) {};
            \node[circle, draw, fill=purple!20, minimum size=0.4cm] (b1) at (3, 0.8) {};
            \node[circle, draw, fill=purple!20, minimum size=0.4cm] (b2) at (4, 0.8) {};
            \draw[gray, thick] (P) -- (A); \draw[gray, thick] (P) -- (B);
            \draw[gray] (A) -- (a1); \draw[gray] (A) -- (a2);
            \draw[gray] (B) -- (b1); \draw[gray] (B) -- (b2);
            \draw[<->, dashed, red, thick] (A) -- node[above, font=\scriptsize] {$D<\tau$} (B);
            \draw[-{Latex[length=3mm]}, ultra thick] (4.5, 2.5) -- (6, 2.5);
            \node[circle, draw, fill=blue!10, minimum size=1cm, font=\small] (P2) at (7.5, 4.5) {P};
            \node[circle, draw, fill=green!40, minimum size=1cm] (AB) at (7.5, 2.5) {AB};
            \node[circle, draw, fill=orange!20, minimum size=0.4cm] (na1) at (6.3, 0.8) {};
            \node[circle, draw, fill=orange!20, minimum size=0.4cm] (na2) at (7.1, 0.8) {};
            \node[circle, draw, fill=purple!20, minimum size=0.4cm] (nb1) at (7.9, 0.8) {};
            \node[circle, draw, fill=purple!20, minimum size=0.4cm] (nb2) at (8.7, 0.8) {};
            \draw[gray, thick] (P2) -- (AB);
            \foreach \x in {na1, na2, nb1, nb2} \draw[gray] (AB) -- (\x);
        \end{tikzpicture}
        \caption{Merge}
        \label{fig:merge}
    \end{subfigure}
    \hfill 
    \begin{subfigure}[b]{0.48\linewidth}
        \centering
        \begin{tikzpicture}[scale=0.38, transform shape]
            \node[circle, draw, fill=blue!10, minimum size=1cm, font=\small] (P) at (2, 5) {P};
            \node[circle, draw, fill=white, line width=1pt, minimum size=1.2cm] (A) at (2, 3) {Mixed};
            \node[circle, draw, fill=orange!20, minimum size=0.4cm] (g1) at (0.6, 1.2) {};
            \node[circle, draw, fill=orange!20, minimum size=0.4cm] (g2) at (1.1, 0.8) {};
            \node[circle, draw, fill=purple!20, minimum size=0.4cm] (g3) at (1.8, 0.6) {};
            \node[circle, draw, fill=purple!20, minimum size=0.4cm] (g4) at (2.4, 0.6) {};
            \node[circle, draw, fill=teal!20, minimum size=0.4cm] (g5) at (3.1, 0.8) {};
            \node[circle, draw, fill=teal!20, minimum size=0.4cm] (g6) at (3.6, 1.2) {};
            \draw[gray, thick] (P) -- (A);
            \foreach \x in {g1, g2, g3, g4, g5, g6} \draw[gray] (A) -- (\x);
            \node[left, font=\scriptsize, red] at (1.2, 3) {High Incoher.}; 
            \draw[-{Latex[length=3mm]}, ultra thick] (4, 3) -- (5.5, 3);
            \node[above, font=\scriptsize] at (3.75, 3) {KMeans};
            \node[circle, draw, fill=blue!10, minimum size=1cm, font=\small] (P2) at (8, 5) {P};
            \node[circle, draw, fill=orange!10, minimum size=0.7cm, font=\scriptsize] (A1) at (6.0, 3) {C1};
            \node[circle, draw, fill=purple!10, minimum size=0.7cm, font=\scriptsize] (A2) at (8.0, 3) {C2};
            \node[circle, draw, fill=teal!10, minimum size=0.7cm, font=\scriptsize] (A3) at (10.0, 3) {C3};
            \node[circle, draw, fill=orange!20, minimum size=0.4cm] (ng1) at (5.6, 1.0) {};
            \node[circle, draw, fill=orange!20, minimum size=0.4cm] (ng2) at (6.4, 1.0) {};
            \node[circle, draw, fill=purple!20, minimum size=0.4cm] (ng3) at (7.6, 1.0) {};
            \node[circle, draw, fill=purple!20, minimum size=0.4cm] (ng4) at (8.4, 1.0) {};
            \node[circle, draw, fill=teal!20, minimum size=0.4cm] (ng5) at (9.6, 1.0) {};
            \node[circle, draw, fill=teal!20, minimum size=0.4cm] (ng6) at (10.4, 1.0) {};
            \draw[gray, thick] (P2) -- (A1); \draw[gray, thick] (P2) -- (A2); \draw[gray, thick] (P2) -- (A3);
            \draw[gray] (A1) -- (ng1); \draw[gray] (A1) -- (ng2);
            \draw[gray] (A2) -- (ng3); \draw[gray] (A2) -- (ng4);
            \draw[gray] (A3) -- (ng5); \draw[gray] (A3) -- (ng6);
        \end{tikzpicture}
        \caption{Split}
        \label{fig:split}
    \end{subfigure}
    
    \caption{Visual illustration of the four atomic operations in SHAPE: (a) Grafting, (b) Pruning, (c) Merge, and (d) Split. These operations dynamically adapt the hierarchical topology.}
    \label{fig:shape}
\end{figure}

\subsection{Partial Data Sharing}

SOFA-FL also allows data sharing between clients to alleviate the problem of data heterogeneity. The ratio of data sharing varies across clusters: clients within the same subgroup may share more data due to inner similarities and for better personalization, while clients in different subgroups may share less data for privacy preservation and global generalization. The adaptive clustered data sharing mechanism will mitigate the impact of heterogeneous data and improve personalization while limiting potential privacy leakage issues in a controlled manner.

The data sharing includes two phases, gathering and distributing. During gathering stage, clients will send partial data to their parent cluster node, same for the upper-level cluster node. Then in the distributing stage, the upper-level cluster will distribute their collected shared data to its children. Finally, clients will add the shared data to their training dataset. 

\begin{figure}[H]
    \centering
    \begin{subfigure}{0.45\linewidth}
        \centering
        \includegraphics[width=\linewidth]{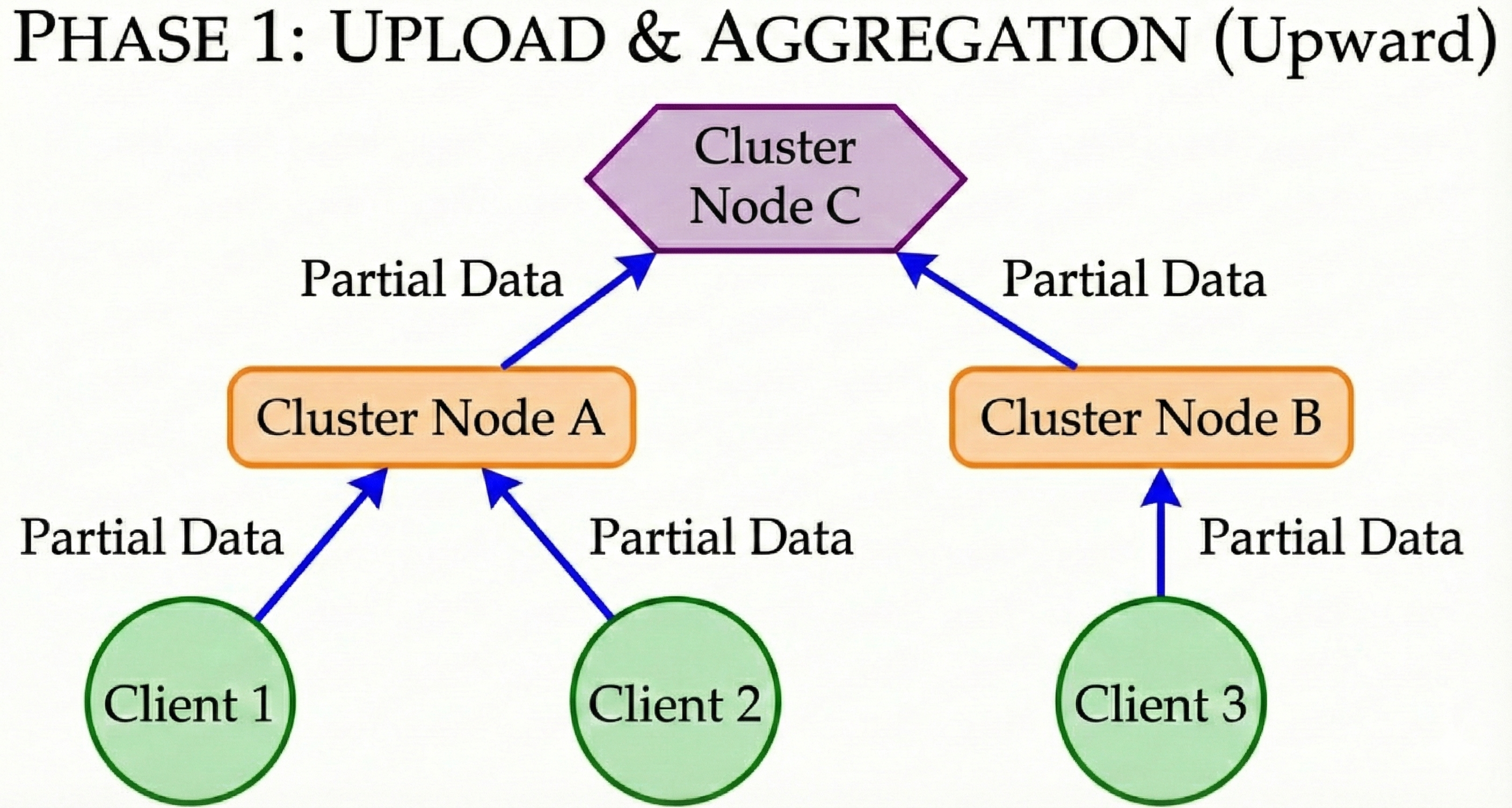}
        \caption{upward}
    \end{subfigure}
    \begin{subfigure}{0.45\linewidth}
        \centering
        \includegraphics[width=\linewidth]{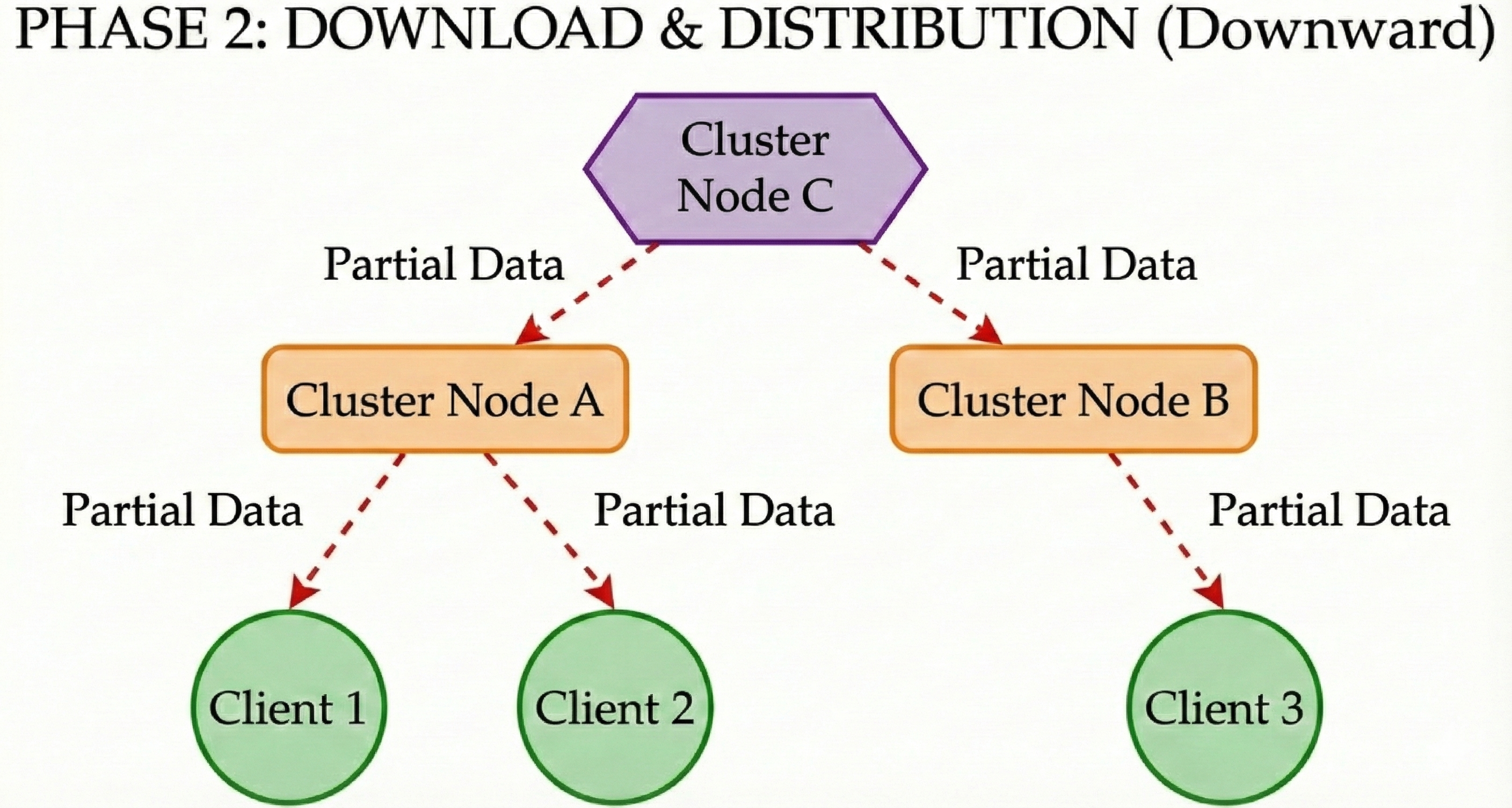}
        \caption{downward}
    \end{subfigure}
\end{figure}

\subsection{Global Objective Function} 
    \begin{align*}
        \min_{\{x_{u}\},S} \mathcal{L} &= \underbrace{\sum_{j\in\mathcal{Q}}w_{j}\sum_{u\in\mathcal{C}_{j}}w_{u}F_{u}(x_{u})}_{\text{\textbf{(1)} Hierarchical Model Loss}} 
        + \underbrace{\sum_{j\in\mathcal{Q}}\alpha_{j}\frac{1}{\mathcal{C}_j}\sum_{u_{1},u_{2}\in\mathcal{C}_{j}}\mathcal{D}(x_{u_{1}},x_{u_{2}})}_{\text{\textbf{(2)} Inter-Cluster Reg.}} 
        + \underbrace{\sum_{j\in\mathcal{Q}}\beta_{j}\sum_{u\in\mathcal{C}_{j}}\mathcal{D}(x_{u},x_{j})}_{\text{\textbf{(3)} Intra-Cluster Reg.}}
    \end{align*}

    \begin{itemize}
        \item[\textbf{(1)}] Minimizes weighted local loss across all clusters.
        \item[\textbf{(2)}] Enforces similarity among sibling nodes in the same cluster.
         \item[\textbf{(3)}] Maintains consistency between child nodes and their parent.
    \end{itemize}

\section{Experiments}

\subsection{Experiment setup and benchmark} 
During our experiments, we used the basic convolutional neural network with 2 layer convolution and 3 fully connected layer to give the prediction output. The dataset we use is MNIST. We distributed the dataset to each client in a heterogeneous manner by applying Dirichlet distribution of $\alpha=1$. The local loss function is cross entropy loss to measure the classification accuracy. The other hyperparameters include 20 clients, 20 communication rounds, 5 local updates, and a data sharing ratio of 0.1.

We used HypCluster model as the benchmark. The experiment settings are the same across two schemes, including clients number, communication rounds and local updates number. Note that for HypCluster, $K$ determines the total number of clusters and when $K=1$, the model will be reduced to FedAvg. In the following session, we will first show the comparison result for optimal $K$ HypClusters and the result for corresponding optimal result with data sharing in terms of accuracy and fairness. Then we will analyze the effect of partial data sharing in the ablation study. 

\subsection{Experiment results and analysis} 
\begin{figure}
    \centering
    \includegraphics[width=0.5\linewidth]{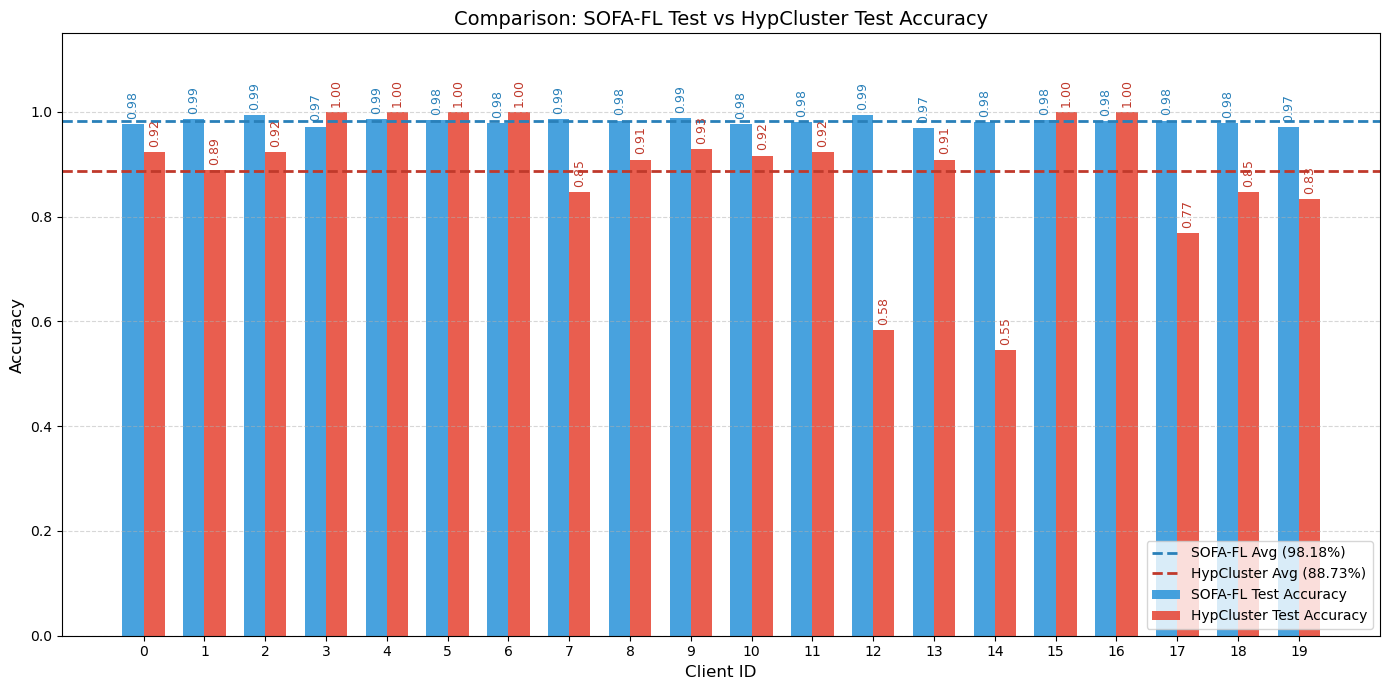}
    \caption{Comparison between SOFA-FL and HypCluster}
    \label{fig:compare}
\end{figure}

We set both the SOFA-FL and HypCluster to have 20 clients and the same data distribution. The optimal results of HypClusters is achieved when the cluster number is 3. We compare the performance of each clients between these two methods in  Fig \ref{fig:compare}. The results clearly show that SOFA-FL outperforms HypCluster across almost all clients. SOFA-FL maintains consistently high accuracy (mostly 0.97–1.00), while HypCluster not only performs lower on average but also shows large drops on several clients. This indicates that SOFA-FL provides stronger generalization and far more stable performance under heterogeneous client data, making it the superior approach overall. 

In addition, table \ref{tab:compare} indicates SOFA-FL shows stronger fairness across clients. Its near-perfect Jain’s Index and tiny performance gap indicate highly uniform accuracy, with even the weakest clients still achieving around 97\%. By contrast, HypCluster suffers large disparities: its accuracy swings widely, and its bottom clients drop to barely above 50\%. This shows that SOFA-FL delivers both higher accuracy and far more equitable treatment for all clients.

\begin{table}[htbp]
  \centering
  \caption{Fairness Performance Comparison: SOFA-FL vs. HypCluster}
  \label{tab:fairness_comparison}
  \begin{tabular}{lcc}
    \toprule
    \textbf{Metric} & \textbf{SOFA-FL (Ours)} & \textbf{HypCluster} \\
    \midrule
    Mean Accuracy & \textbf{98.18\%} & 88.73\% \\
    Standard Deviation & \textbf{0.0067} & 0.1251 \\
    Min Accuracy (Worst Client) & \textbf{96.86\%} & 54.55\% \\
    Accuracy Gap (Max - Min) & \textbf{2.57\%} & 45.45\% \\
    Jain's Fairness Index & \textbf{0.9999} & 0.9805 \\
    Avg. Accuracy of Bottom 10\% & \textbf{96.99\%} & 56.44\% \\
    \bottomrule
  \end{tabular}
  \label{tab:compare}
\end{table}

\subsection{Ablation Study}
The effect of data sharing is shown clearly in the local loss images in the appendix section. When we add partial data sharing, the local loss will have a surge in the beginning of local update. Because we use random portion of the dataset, new data will be shared in each round, resulting in increase of local loss and then the model will gradually decrease the loss until next round. 

If we use a fixed random seed, then the local loss will be similar to the case without partial data sharing where the local loss would only increase when the structure is modified.

The test accuracy also increases if we add partial data sharing with fixed random seed. 
\begin{table}[h!]
\centering
\begin{tabular}{|c|c|c|}
\hline
 & client average test accuracy & total average test accuracy \\ \hline
without partial data sharing & 0.9794 & 0.9451 \\ \hline
partial data sharing ratio $0.1$ & 0.9866 & 0.9604 \\ \hline
partial data sharing ratio $0.1$ fixed& 0.9827 & 0.9578 \\ \hline
partial data sharing ratio $0.2$ fixed& 0.9846 & 0.9430 \\ \hline
\end{tabular}
\caption{Test accuracy for ablation study}
\end{table}

\vspace{-1cm}
\section{Conclusion}
In this paper, we proposed a feasible and effective paradigm for clustered federated learning. In federated learning, two main challenges are data drift in evolving environments and data heterogeneity. To address these issues, we introduce a multi-level hierarchical framework with the SHAPE algorithm for adapting to new data patterns, as well as a novel model-based data sharing mechanism to mitigate data heterogeneity.

\newpage 
\bibliographystyle{plainnat}

\bibliography{references}



\newpage 

\appendix 
\section{Appendix}
\subsection{Algorithm} 

\begin{algorithm}[H]
\caption{Dynamic Multi-branch Agglomerative Clustering}
\label{alg:DMAC}
\begin{algorithmic}[1]
\State Initialize node list $[n_i]$
\State Initialize increasing factor $\gamma$
\While{node list has more than 1 node}
    \For{each node $n_i$}
        \For{each node $n_j$}
            \State Calculate distance $D(n_i,n_j)$
        \EndFor
    \EndFor
    \State $\tau \leftarrow min D(n_i,n_j)$
    \For{each pair $(n_u,n_v)$ that $D(n_{ui},n_{vi})\leq\tau * \gamma$}
        \State $[g_u]\leftarrow \text{clusters of }n_v$
        \State $[g_v]\leftarrow \text{clusters of }n_v$
        \State $g_u \leftarrow \text{Aggregate}(g_u,g_v)$
        \State Add nodes of $g_v$ as children of $g_u$
        \If{$n_u$ in node list }
        \State Remove $n_u$ from node list
        \EndIf
        \If{$n_v$ in node list }
        \State Remove $n_v$ from node list
        \EndIf
    \EndFor
    
    \State Append $[g_i]$ to node list $[n_i]$
\EndWhile

\end{algorithmic}
\end{algorithm}

\begin{algorithm}[H]
\caption{Grafting (Re-Parenting)}
\label{alg:graft}
\begin{algorithmic}[1]  
\Require Child node $n_c$, current parent $n_{\text{pre}}$, candidate parents $\mathcal{C}$, tolerance $\epsilon$
\Ensure Parent of $n_c$ is updated if a significantly closer parent exists

\State $n_{\min} \gets \arg\min_{n \in \mathcal{C}} d(n, n_c)$
\State $d_{\text{pre}} \gets d(n_{\text{pre}}, n_c)$
\State $d_{\min} \gets d(n_{\min}, n_c)$

\If{$d_{\min} (1 + \epsilon) < d_{\text{pre}}$}
    \State Remove edge $(n_{\text{pre}} \rightarrow n_c)$
    \State Add edge $(n_{\min} \rightarrow n_c)$
\EndIf

\end{algorithmic}
\end{algorithm}

\begin{algorithm}[H]
\caption{Pruning (Trim)}
\label{alg:prune}
\begin{algorithmic}[1]
\Require Intermediate node $P$ with parent $G$ and children $\mathcal{C}(P)$
\Ensure Remove redundant pass-through nodes and locally flatten the hierarchy

\If{$|\mathcal{C}(P)| = 1$}
    \State Let $C$ be the unique child in $\mathcal{C}(P)$
    \State Remove edge $(P \rightarrow C)$
    \State Add edge $(G \rightarrow C)$
    \State Remove node $P$

    \EndIf
\end{algorithmic}
\end{algorithm}

\begin{algorithm}[H]
\caption{Consolidation (Merge)}
\label{alg:merge}
\begin{algorithmic}[1]
\Require Sibling nodes $n_a$ and $n_b$ with common parent $P$, merge threshold $\tau$
\Ensure Merge highly similar siblings into a single node

\If{$D(n_a, n_b) < \tau$}
    \State $n_{ab} \gets \text{CreateNode}()$  \Comment{new merged node}
    \State $\mathcal{C}(n_{ab}) \gets \mathcal{C}(n_a) \cup \mathcal{C}(n_b)$

    \ForAll{$C \in \mathcal{C}(n_{ab})$}
        \State Remove edge $(n_a \rightarrow C)$ if it exists
        \State Remove edge $(n_b \rightarrow C)$ if it exists
        \State Add edge $(n_{ab} \rightarrow C)$
    \EndFor

    \State Remove edge $(P \rightarrow n_a)$ and $(P \rightarrow n_b)$
    \State Remove nodes $n_a$ and $n_b$
    \State Add edge $(P \rightarrow n_{ab})$
\EndIf
\end{algorithmic}
\end{algorithm}

\begin{algorithm}[H]
\caption{Purification (Split)}
\label{alg:split}
\begin{algorithmic}[1]
\Require Node $n$ with successors $\text{Succ}(n)$, split threshold $\theta_{\text{split}}$
\Ensure Replace $n$ by several more coherent child clusters if possible

\For{$k = 2$ \textbf{to} $|\text{Succ}(n)|$}
    \State Run KMeans on $\text{Succ}(n)$ to obtain $k$ groups $G_1, \dots, G_k$
    \State $\mathcal{N}_{\text{new}} \gets \emptyset$
    \For{$j = 1$ \textbf{to} $k$}
        \State $n_j \gets \text{CreateNode}()$ with children $G_j$
        \State $\mathcal{N}_{\text{new}} \gets \mathcal{N}_{\text{new}} \cup \{n_j\}$
    \EndFor

    \If{$\forall n_j \in \mathcal{N}_{\text{new}}:\ \text{Incoherence}(n_j) \leq \theta_{\text{split}}$}
        \State Let $P$ be the parent of $n$
        \State Remove edge $(P \rightarrow n)$
        \ForAll{$n_j \in \mathcal{N}_{\text{new}}$}
            \State Add edge $(P \rightarrow n_j)$
        \EndFor
        \State Remove node $n$
        \State \Return  \Comment{split accepted}
    \EndIf
\EndFor
\State \Return  \Comment{no valid split found}\end{algorithmic}
\end{algorithm}

\appendix
\label{appendix:shape_algorithm}
\begin{algorithm}[H]
\caption{Partial Data Sharing}
\label{alg:partial_data_sharing}
\begin{algorithmic}[1]

\State \textbf{Initialize} clients and clusters $n_i$ with dataset $d_i$
\State \textbf{Initialize} data sharing ratio $\alpha$

\Statex
\Function{Gather}{$n_r$} \Comment{Gathering Phase}
    \For{each child node $n_i$ of $n_r$}
        \State \Call{Gather}{$n_i$}
        \State Add random subset $\alpha \cdot d_i$ to root shared dataset $d_r$
    \EndFor
\EndFunction

\Statex
\Function{Distribute}{$n_r$} \Comment{Distributing Phase}
    \For{each child node $n_i$ of $n_r$}
        \State Add root's dataset $d_r$ to child dataset $d_i$
        \State \Call{Distribute}{$n_i$}
    \EndFor
\EndFunction

\end{algorithmic}
\end{algorithm}

\begin{algorithm}[H]
\caption{Partial Data Sharing}
\label{alg:DMAC}
\begin{algorithmic}[1]
\State Initialize clients and clusters $n_i$ with their dataset $d_i$
\State Initialize data sharing ratio $\alpha$ 
\State Start from root cluster. 
\For{each child node $n_i$ of root}
    \State Recursively gather shared data from $n_i$. 
    \State Add random subdataset $\alpha * d_i$ to root's shared dataset $d_r$. 
\EndFor

\State Start from root cluster. 
\For{each child node $n_i$ of root}
    \State Add root's dataset $d_r$ to child node $d_i$. 
    \State Recursively distribute shared data to child $n_i$.
\EndFor

\end{algorithmic}
\end{algorithm}

\begin{algorithm}[H]
\caption{SOFA-FL}
\label{alg:DMAC}
\begin{algorithmic}[1]
\State Initialize clients $n_i$ and assign non-IID data to each client. 
\State Apply DMAC to construct initial clustering structure. 
\For{each communication round $r$}
    \State Do local update. 
    \State Update cluster model and compare the loss. 
    \State Apply SHAPE to restructure the cluster. 
    \State Apply Partial Data Sharing to update shared data. 
\EndFor
\end{algorithmic}
\end{algorithm}

\subsection{Experiment Results}
\begin{figure}[H]
    \centering
        \centering
        \includegraphics[width=0.9\linewidth]{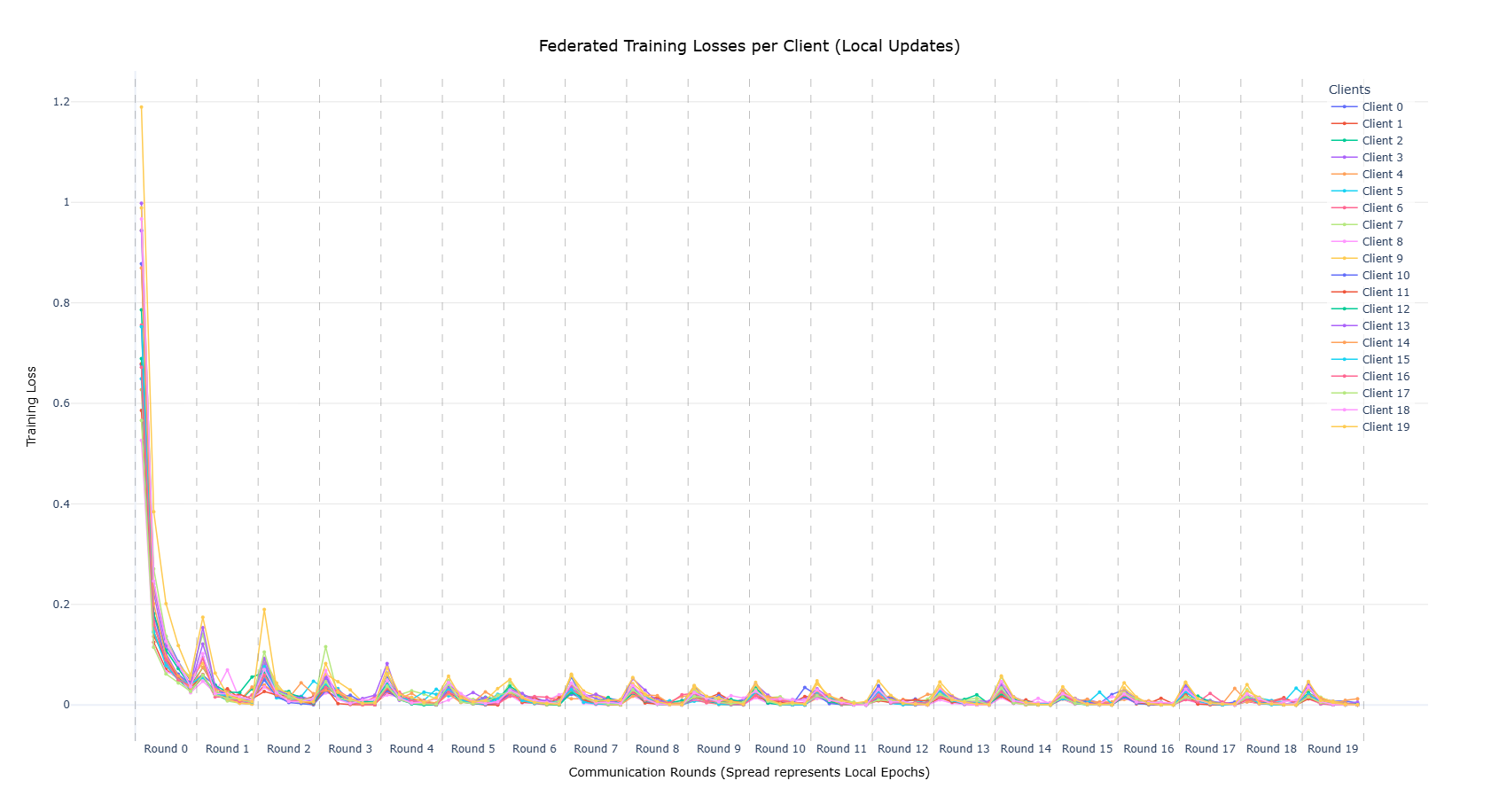}
        \caption{with partial data sharing}
\end{figure}

\begin{figure}[H]
    \centering
        \centering
        \includegraphics[width=0.9\linewidth]{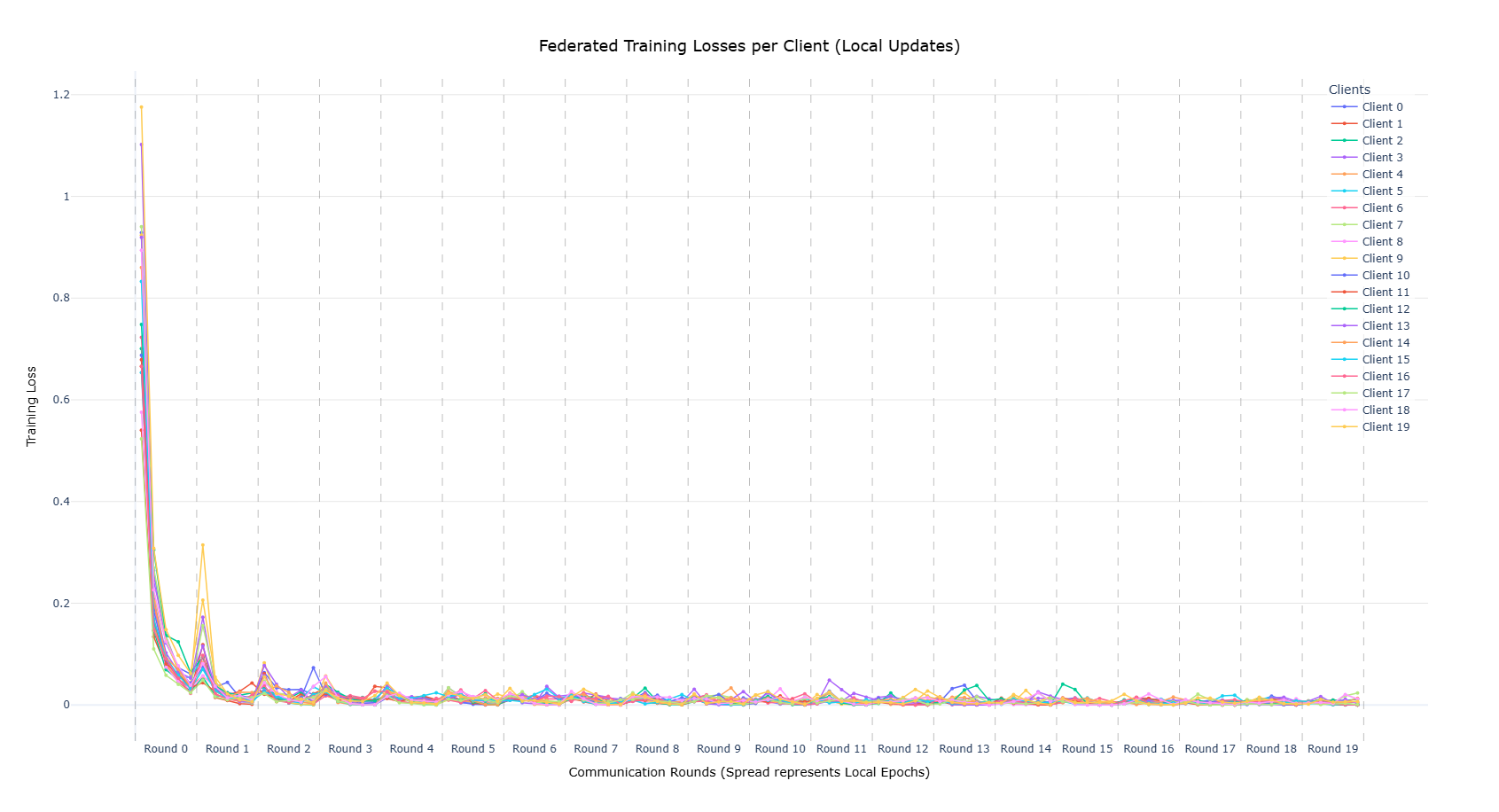}
        \caption{without partial data sharing}
\end{figure}

\begin{figure}[H]
    \centering
        \centering
        \includegraphics[width=0.9\linewidth]{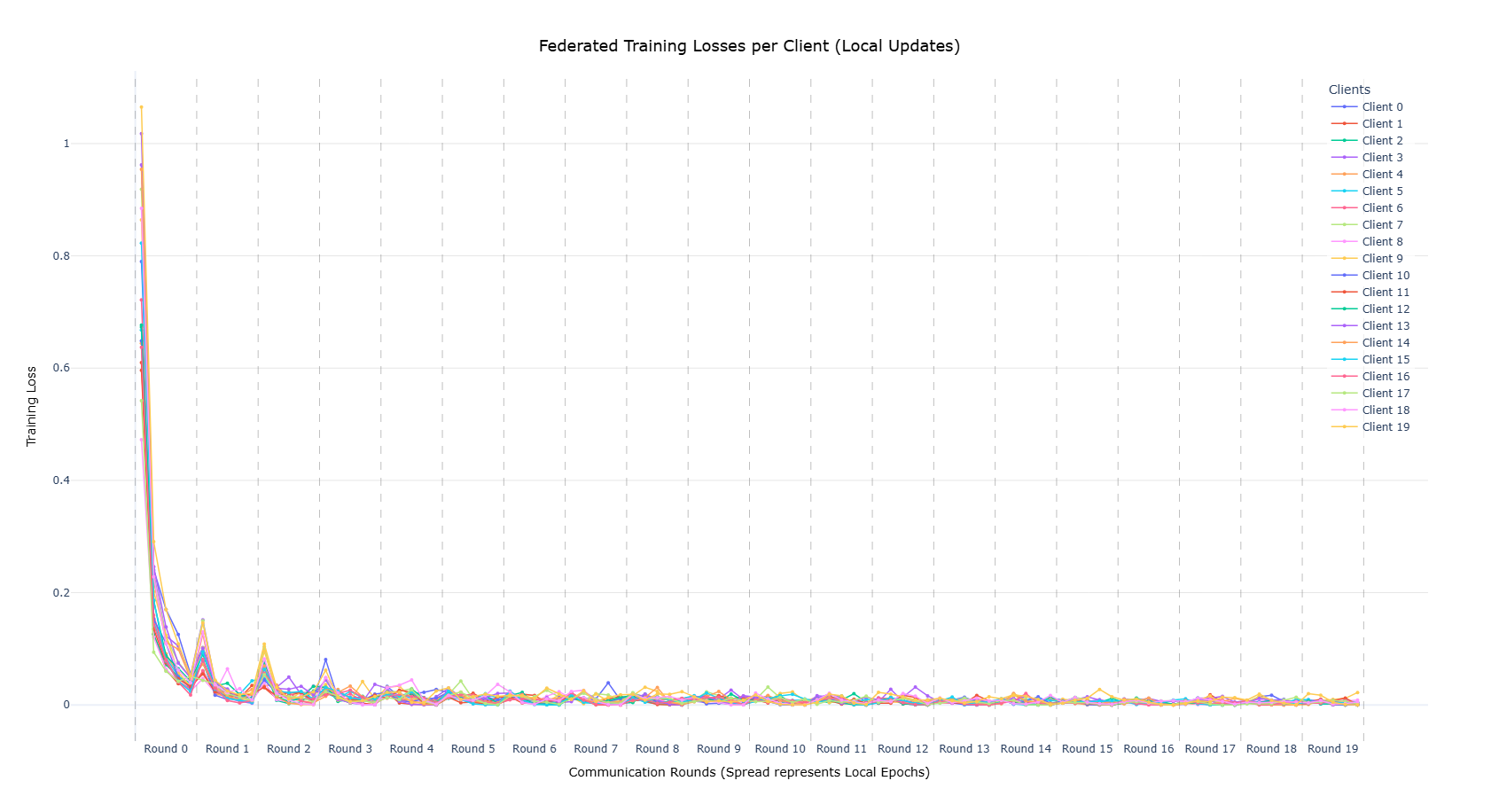}
        \caption{partial data sharing fixed with ratio $0.1$}
\end{figure}

\begin{figure}[H]
    \centering
        \centering
        \includegraphics[width=0.9\linewidth]{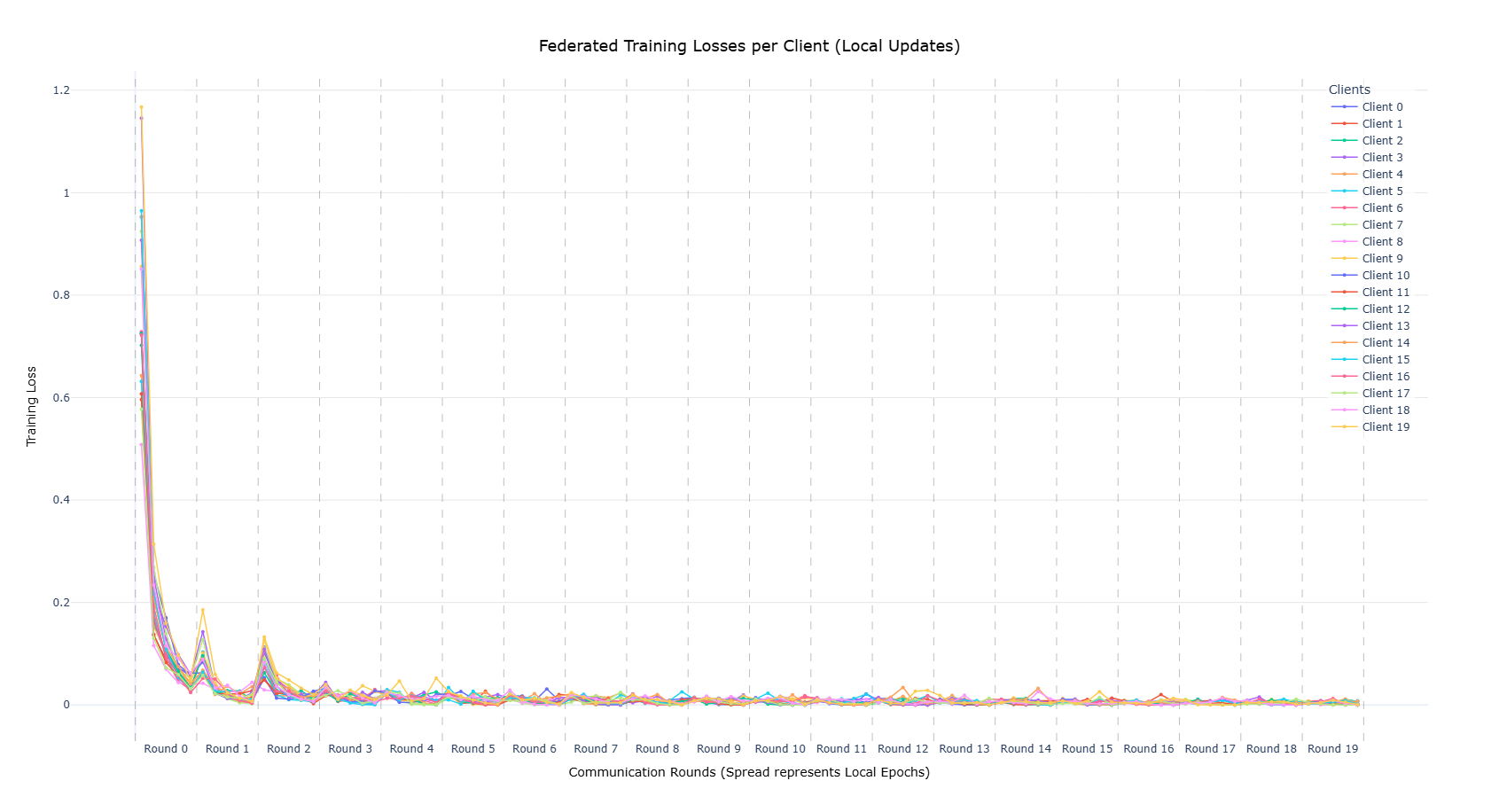}
        \caption{partial data sharing fixed with ratio $0.2$}
\end{figure}



\end{document}